\documentclass[11pt]{article}
\usepackage[preprint]{acl}
\usepackage{times}
\usepackage{latexsym}
\usepackage[T1]{fontenc}
\usepackage[utf8]{inputenc}
\usepackage{microtype}
\usepackage{graphicx}
\usepackage{amsmath}
\usepackage{amssymb}
\usepackage{booktabs}
\usepackage{capt-of}

\title{Geometric Structure Predicts Per-Example Neighborhood Loss Under PCA Truncation}

\author{Sriram Loganathan, Gokul Anand, Ragavan Arivazhagan \\
  Department of Computer Science \\
  San Jose State University \\
  San Jose, CA, USA}

\begin{document}
\maketitle

\begin{abstract}
Dimensionality reduction can preserve the average performance of text
embeddings even when a substantial fraction of their dimensions is
removed. Aggregate metrics, however, do not reveal whether degradation
is distributed uniformly across examples. We study per-example
neighborhood stability under PCA truncation across static word
embeddings and modern sentence encoders. Per-item neighbor retention
varies widely and is predictable from geometric properties of the
representation and its alignment with the retained PCA subspace. The
fraction of centered-vector energy retained by the truncated subspace
is strongest for CBOW and RI, with neighborhood contrast leading on
SkipGram and the modern/Matryoshka settings (GTE approximately tied), while local density is associated with retention in marginal analysis but adds little unique information once energy and
contrast are controlled (R-squared increments 0.001 to 0.010). Local intrinsic dimensionality accounts for
more variation than density on modern sentence encoders in pairwise
comparison. On SciFact, query density predicts retrieval
degradation under aggressive compression; on NFCorpus the primary
32d association does not replicate, and a direct variance-restriction
test leaves that null
unresolved. Average compression performance conceals systematic
per-example variation that geometry makes visible.
\end{abstract}

\section{Introduction}
\label{sec:intro}

Truncation is routine for serving retrieval and classification, and published means change little for similarity, analogies, and retrieval. A stable mean does not identify which representations are damaged, since the same average is consistent with concentrated loss in sparse or unusual neighborhoods.

We ask what properties of full-space representations predict per-example damage before compressing. We measure per-item neighbor retention under PCA truncation across static word vectors, modern sentence encoders at 384 and 768 dimensions, native Matryoshka prefixes, and gold relevance judgments on two retrieval benchmarks. The candidate predictors span local density, retained subspace energy, neighborhood contrast, boundary gap, and intrinsic dimensionality, tested against frequency, hubness, margin, and norm controls.

No single predictor dominates universally. Retained subspace energy leads on CBOW and RI while neighborhood contrast leads on SkipGram and the modern and Matryoshka encoders (GTE approximately tied); local density is consistently associated with retention but contributes little unique information jointly. On SciFact, query density predicts gold retrieval degradation at 32d and 64d; the primary 32d association does not replicate on NFCorpus, and a direct variance-restriction test leaves that null unresolved.

Contributions: (1) A per-example measurement of compression fragility across static, modern, and Matryoshka representations. (2) A joint decomposition showing that the informative geometric signals differ by representation regime, with retained energy and neighborhood contrast explaining most of density's marginal association. (3) Downstream retrieval evidence on SciFact and mixed transfer to NFCorpus, together with limits on adaptive allocation.

\section{Related Work}
\label{sec:related}

\paragraph{Embedding compression and PCA.}
PCA truncation with post-processing is evaluated on aggregate
similarity benchmarks \cite{raunak-etal-2019-effective}. Quantized
and distilled variants report means over translation and language
modeling tasks. Recent large-scale studies establish that random
dimension deletion barely moves aggregate retrieval and
classification scores \cite{takeshita-etal-2025-randomly}, and that
task-conditioned redundancy varies by task type
\cite{tsukagoshi-sasano-2025-redundancy}. Takeshita et al.\ (2025) report aggregate robustness curves and anisotropy analyses, but do not examine per-example stratification. Follow-up work shows truncation robustness is largely
inherent rather than Matryoshka-induced except under heavy reduction
\cite{takeshita-etal-2026-mrl}, and that PCA compression can help or
hurt retrieval by domain \cite{zuo-khashabi-2026-efficiency}.
These studies report aggregate robustness without testing whether
full-space geometry predicts per-example neighborhood loss under
truncation.

\paragraph{Variable and nested dimensions.}
Matryoshka representation learning trains encoders whose prefixes
work standalone \cite{kusupati-etal-2022-matryoshka}. Adaptive
per-word code lengths are learned end-to-end by downstream tasks
\cite{kim-etal-2020-adaptive}. Frequency-aware sparse coding
compresses rare-token embeddings of fine-tuned models
\cite{tamura-etal-2024-compressing}. These approaches operate during training or use frequency-based allocation.

\paragraph{Representation geometry and anisotropy.}
Static embeddings are anisotropic and a few rogue dimensions dominate
cosine \cite{ethayarajh-2019-contextual,timkey-van-schijndel-2021-bark}.
Removing dominant directions changes aggregate behavior
\cite{mu-etal-2018-all}. Modern work links PC removal to metric
choice \cite{parupudi-2026-anisotropy} and shows anisotropy alone
does not explain semantic deficits \cite{fuster-baggetto-fresno-2022-anisotropy}.
Dimensionality-reduction quality is itself studied at the aggregate
level through trustworthiness and continuity
\cite{venna-kaski-2001-trustworthiness} and the co-ranking framework
\cite{lee-verleysen-2009-coranking}, which score whole projections
rather than predicting per-example retention from full-space features.
Global intrinsic dimension of network representations is estimated
with TwoNN across layers \cite{ansuini-etal-2019-intrinsic}, again as
a population summary rather than a per-item predictor.

\paragraph{Per-item stability and heterogeneity.}
Word-embedding stability across retrains varies per word, with
frequency playing a complex role (high-frequency stable,
low-frequency unstable, medium-frequency high-variance)
\cite{wendlandt-etal-2018-factors,hellrich-hahn-2016-bad}.
Frequency distorts similarity judgments and rare words segregate
into static subregions \cite{gong-etal-2018-frage}.
These establish per-item heterogeneity under retraining, not under PCA truncation at inference time. Nearest-neighbor difficulty is also studied per query: local intrinsic dimensionality serves as a hardness signal for approximate search \cite{amsaleg-etal-2015-lid}, but that literature does not test which geometry predicts neighbor loss caused by truncating the embeddings themselves.

\paragraph{Frequency and rare-word behavior.}
Top principal components track frequency bins under shuffled-corpus
controls \cite{valentini-etal-2023-investigating}. We treat
frequency as a confound throughout (log controls and within-decile
analyses) rather than as the predictor.

Prior work has largely evaluated compression at the aggregate level or studied per-item stability under different perturbations. We instead ask which full-space geometric properties predict per-example neighborhood loss under truncation.

\section{Measuring Per-Example Compression Damage}
\label{sec:method}

Let $V \in \mathbb{R}^{N \times d}$ be row-normalized embeddings of
$N$ items (words or sentences); all similarities below are dot
products on L2-normalized vectors (cosine). For PCA truncation to
$p < d$ on static families we center with the full-vocabulary mean
and keep top-$p$ SVD components, then renormalize; on modern
encoders we center with a train-split mean (SST-2 train subset,
SciFact corpus) and keep top-$p$ components, then renormalize. For item $i$, let $A_i$ be its top-$k$ neighbor set at
full rank and $B_i(p)$ at rank $p$, self excluded, $k=10$
throughout. Retention is $R_{10}(i,p) = |A_i \cap B_i(p)|/10$;
loss is $L_{10} = 1 - R_{10}$. Local density is
$D_{10}(i)$, the mean cosine to the top-10 neighbors in full space.
Competing signals: hubness $N_{10}$ (times $i$ appears in others'
top-10), margin (top-1 minus top-2 similarity), vector norm, local
intrinsic dimension (Levina-Bickel MLE over the 50-neighborhood
\cite{levina-bickel-2004-maximum}, with TwoNN background
\cite{facco-etal-2017-twonn}; see also \cite{ruppik-etal-2025-less,valeriani-etal-2023-geometry,lee-etal-2025-shared}),
log frequency or a rarity proxy. For sentences, rarity is the mean over lowercased whitespace tokens of negative log frequency plus one, with frequencies counted on the evaluation sentences. Associations are Pearson partial correlations with
paired bootstrap 95\% CIs (2,000 resamples for static embeddings and 5,000 for modern encoders).
Neighbor sets use a chunked exact top-$k$ implementation with
self-exclusion by id, validated against brute-force search. Replication criteria were specified before evaluating the modern encoders; details are provided in Appendix~\ref{sec:appendix-repro}.
Joint analysis uses one primary matched-severity width per family
(static 200d to 32d; 384d to 64d; 768d to 128d; Nomic native and PCA
128d separately). Competitors are computed from full-space rank-ordered
similarities to rank 50: boundary gap $s_{10}-s_{11}$, neighborhood
contrast (mean $s_{1:10}$ minus mean $s_{11:30}$), retained energy
$\|P_px_c\|^2/\|x_c\|^2$ on centered vectors with the PCA operator used for truncation
(and a separately labeled native-prefix analogue for Nomic), and
Levina-Bickel ID over the 50-neighborhood. We call a partial correlation marginal when it controls frequency only, pairwise when it additionally controls one competing geometric predictor, and joint when it conditions on the full predictor set. Joint partial correlations condition on all predictors simultaneously. To quantify density's incremental predictive contribution, we additionally report the change in $R^2$ when density is added after the competing variables. Decoupled variants use ranks 11--30
and 11--50 only. Uncertainty uses paired bootstrap CIs; confirmatory tests are
Holm-corrected.

\section{Experimental Setup}
\label{sec:setup}

Static families: Word2Vec CBOW 200d window 4 on fairytales (18,210
words) \cite{mikolov-etal-2013-efficient}; Skip-Gram 200d same
corpus (17,620 words); Random Indexing with damped PPMI graph
averaging on text8sub (18,254 words). Per family $n=1500$ sampled
words (all eval-covered words plus random fill, seed 7).
Truncations 100/50/32. Eval: WordSim353 (231 covered pairs)
\cite{finkelstein-etal-2002-placing} and a 40-question analogy
subset. Modern encoders: MiniLM-L6 (22.7M, 384d)
\cite{wang-etal-2020-minilm}, GTE-small (33M, 384d)
\cite{li-etal-2023-gte}, MPNet-base (109M, 768d)
\cite{song-etal-2020-mpnet}, Nomic v1.5 (137M, 768d, Matryoshka-trained)
\cite{nussbaum-etal-2024-nomic}. Each Nomic input is prefixed with the literal string \texttt{classification:}. Native truncation then applies layer norm
to the full 768 dimensions, slices the first $K$ dimensions, and L2
renormalizes. Sentence eval: SST-2 validation (872 sentences)
\cite{socher-etal-2013-recursive} with
PCA fit on an 8000-sentence train subset (seed 7); truncations matched
by severity ratio (384d: 128/64/32; 768d: 256/128/64). The severity
analysis adds a PCA-256 anchor on MiniLM only (256/128/64/32); the 256d
point is not severity-matched. Retrieval:
SciFact (5,183 docs, 300 queries, test qrels) and NFCorpus (3,633
docs, 323 queries, test qrels) \cite{thakur-etal-2021-beir}; PCA fit
on corpus only. SciFact uses MiniLM-L6 and NFCorpus uses GTE-small. Query density is measured against the corpus index.
We report two retrieval analyses: gold relevance
changes against test qrels ($\Delta$ = truncated minus full, so
negative values indicate degradation), and label-free neighborhood
overlap losses (1 minus overlap, so positive values indicate
degradation). Gold metrics use 300 qrel-matched queries; label-free
overlap uses the first 300 cached queries. All runs are CPU-only with no external APIs.

\section{Local Geometry Predicts Compression Fragility}
\label{sec:static}

\begin{figure}[t]
\centering
\includegraphics[width=\columnwidth]{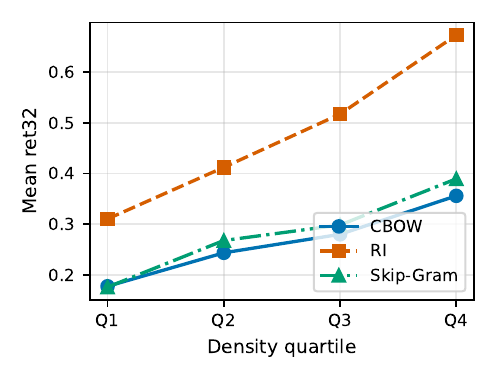}
\caption{Dose response: ret32 by density quartile on three static families. Monotone on all families.}
\label{fig:quartile}
\end{figure}

Mean retention at 32d varies across words on every family (Table~\ref{tab:static} for means 0.264, 0.478, and 0.283; per-word 32d SD near 0.2 from the per-row distributions). Controlling for log frequency, the density partials are 0.498 CI [0.454, 0.537] for CBOW, 0.666 CI [0.636, 0.693] for RI, and 0.541 CI [0.504, 0.580] for SkipGram. After controlling for
density, hubness partials fall to 0.027, 0.040, and 0.020. On CBOW,
the norm partial falls from 0.42 (raw) to 0.13 controlling log
frequency (RI and SG norm partials are already near zero at -0.065
and -0.081). Margin partials lie between -0.171 and +0.171 on all
families. Retention rises monotonically across density quartiles
on all families (Figure~\ref{fig:quartile}).

\begin{table}[t]
\centering
\begin{tabular}{lccc}
\toprule
Family & ret100 & ret50 & ret32 \\
\midrule
CBOW/fairytales & 0.616 & 0.378 & 0.264 \\
RI/text8sub & 0.819 & 0.620 & 0.478 \\
SkipGram/fairytales & 0.591 & 0.398 & 0.283 \\
\bottomrule
\end{tabular}
\caption{Mean neighbor retention @10 by PCA truncation ($n=1500$ each).}
\label{tab:static}
\end{table}

\begin{table}[t]
\centering
\small
\setlength{\tabcolsep}{4pt}
\begin{tabular}{lccccc}
\toprule
Family & density & hub. & hub.$\mid$den. & norm & margin \\
\midrule
CBOW & 0.498 & 0.398 & 0.027 & 0.128 & 0.091 \\
RI & 0.666 & 0.473 & 0.040 & -0.065 & -0.171 \\
SkipGram & 0.541 & 0.377 & 0.020 & -0.081 & 0.171 \\
\bottomrule
\end{tabular}
\caption{Partial correlations with ret32 controlling log frequency, plus hubness controlling density and log frequency. Raw hubness predicts retention, but the association falls near zero after controlling for density and log frequency in all families. Dual control applies to the hub|den column only; density, hubness, norm, and margin columns control log frequency.}
\label{tab:partials}
\end{table}

\section{Generalization to Modern and Matryoshka Encoders}
\label{sec:modern}

\begin{figure}[t]
\centering
\includegraphics[width=\columnwidth]{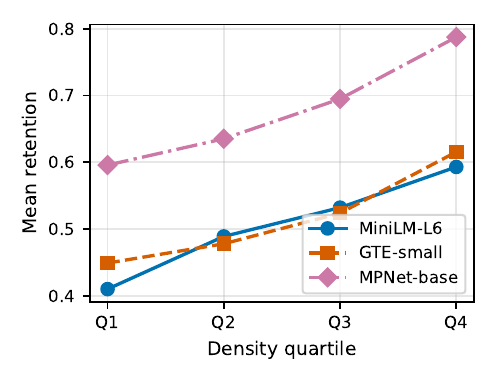}
\caption{Modern dose response by density quartile at matched severity (ret64 for 384d MiniLM-L6 and GTE-small, ret128 for 768d MPNet-base), SST-2 val, $n=872$. Monotone on all three.}
\label{fig:modern}
\end{figure}

On the two 384d encoders at 64d, density partials are 0.416 CI [0.360, 0.473] for MiniLM with hubness at 0.011, and 0.364 CI [0.306, 0.420] for GTE-small with hubness at -0.066 (CI [-0.136, +0.009]). The hubness estimate is near zero for MiniLM; for GTE-small, the confidence interval is wider and extends farther from zero.
On MPNet-base 768d, density partial is 0.517 CI [0.473, 0.561] at 64d
with hubness at 0.032 controlling density; at the matched-severity width of 128d it is 0.485 CI [0.433, 0.534]. Quartile response is monotone on all three tested encoders (Figure~\ref{fig:modern}).

On Nomic v1.5, mean retention is higher under native truncation at each tested width (0.709/0.598/0.473
vs 0.532/0.487/0.424 at 256/128/64); the paired difference of density partials at 128d is $-0.035$ CI $[-0.100, +0.031]$, which does not distinguish the two operators (no paired retention-difference CI was computed). Density remains associated with retention under both operators, with partial correlations of 0.383/0.395/0.428 native and 0.420/0.430/0.389 PCA. SST-2 probe flips by density are rare (up to 7.2\% MiniLM,
1.7\% GTE). Section 8 compares these signals jointly.
\section{Downstream Retrieval Consequences}
\label{sec:downstream}

On SciFact with test qrels ($n=300$), mean $\Delta$nDCG@10
(truncated minus full) is $-0.034$, $-0.105$, $-0.192$ and mean
$\Delta$recall is $-0.026$, $-0.092$,
$-0.186$ at 128, 64, 32 dimensions; negative values indicate
degradation. At 32d, density partials reach
$+0.249$ for nDCG (CI 0.129 to 0.354) and $+0.266$ for recall (CI
0.166 to 0.362); positive sign means denser queries lose less
quality (full per-width estimates in Table~\ref{tab:tiera}). Without labels,
neighborhood overlap loss, defined as 1 minus overlap so larger means worse, rises from 0.222 to 0.514 across cutoffs.
Density correlates positively with neighborhood retention on this label-free scale, with partials $r$ 0.543/0.526/0.530 at 128/64/32 (equivalently, denser queries lose less overlap). Note this sign convention differs from gold $\Delta$ in Table~\ref{tab:tiera}, where negative means worse.

\begin{table}[t]
\centering
\begin{tabular}{lccc}
\toprule
Dim & mean $\Delta$ & partial & 95\% CI \\
\midrule
\multicolumn{4}{c}{\textit{$\Delta$nDCG@10}} \\
128 & $-0.034$ & $+0.098$ & $[-0.037, 0.227]$ \\
64 & $-0.105$ & $+0.142$ & $[0.025, 0.252]$ \\
32 & $-0.192$ & $+0.249$ & $[0.129, 0.354]$ \\
\multicolumn{4}{c}{\textit{$\Delta$Recall@10}} \\
128 & $-0.026$ & $+0.174$ & $[0.077, 0.267]$ \\
64 & $-0.092$ & $+0.157$ & $[0.051, 0.255]$ \\
32 & $-0.186$ & $+0.266$ & $[0.166, 0.362]$ \\
\bottomrule
\end{tabular}
\caption{SciFact gold changes under truncation ($n=300$). $\Delta$ = truncated minus full, so negative values indicate degradation. Denser queries lose less; all cells except 128d nDCG exclude zero.}
\label{tab:tiera}
\end{table}

The NFCorpus control ($n=323$, GTE-small) gives null 32d partials
$-0.028$, CI $[-0.128, +0.073]$ (nDCG) and $+0.019$, CI $[-0.077, +0.109]$
(recall). The 128d nDCG estimate is $+0.096$ CI $+0.009$ to $+0.180$, while the primary 32d estimates on both metrics include zero. NFCorpus query density
has mean 0.84 std 0.03 (range 0.77--0.94) versus SciFact mean 0.50
std 0.09 (range 0.21--0.76), with smaller degradation overall. NFCorpus exhibits substantially less variation in query density within its own embedding space. Because the two benchmarks use different encoders, their absolute density levels are not directly comparable. Restricting SciFact to narrower density bands attenuates the point estimates monotonically, but the attenuation confidence interval includes zero, leaving range restriction unresolved. The retrieval consequence is supported on SciFact where density varies; NFCorpus does not show a consistent association across metrics and truncation levels. WordSim stratum differences are likewise null
at $n=231$.

\begin{figure}[t]
\centering
\includegraphics[width=\columnwidth]{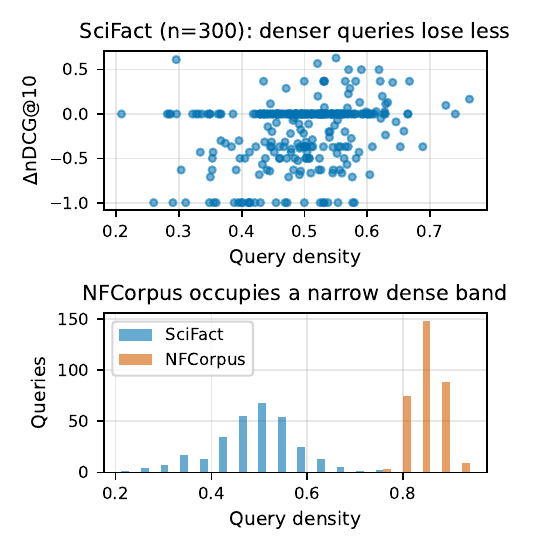}
\caption{Left: per-query $\Delta$nDCG@10 (truncated minus full; negative values indicate degradation) vs density on SciFact at 32d. Right: query density histograms (SciFact with MiniLM-L6, NFCorpus with GTE-small, so absolute density scales are not directly comparable); NFCorpus occupies a narrow dense band with little variation to predict from. The left panel uses the same $\Delta$ convention as Table 3.}
\label{fig:retrieval}
\end{figure}

\section{Joint Geometric Predictors of Retention}
\label{sec:mechanism}

\subsection{Which geometric signals remain informative jointly?}

\begin{figure}[t]
\centering
\includegraphics[width=\columnwidth]{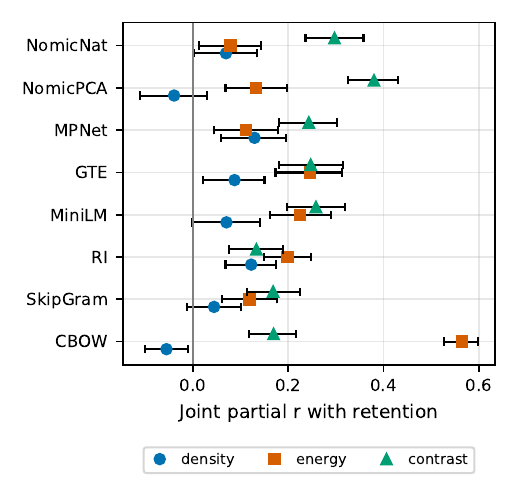}
\caption{Joint multivariable model partials with 95\% CIs: density (circles) against retained energy (squares) and neighborhood contrast (triangles) across eight settings. Energy and contrast are positive throughout; density partials are small (-0.06 to 0.13).}
\label{fig:joint}
\end{figure}

Tested jointly at matched severity, retained energy is positively associated with retention in all eight settings (0.078 to 0.565), the leading joint predictor within CBOW and
RI (for Nomic native, the analogous retained prefix-energy fraction). Neighborhood contrast is also positively associated with retention in all eight (0.133 to
0.380), leading or tying on the other six including SkipGram and a
tie with energy on GTE-small (0.25 versus 0.25). After joint control, density partials are small across all settings (between -0.06 and 0.13), with R-squared increments of
0.001 to 0.010; RI (0.122) and MPNet (0.129) remain Holm-significant
but small. The rank-10 boundary gap is near zero throughout and
the top-1-minus-top-2 margin shows no consistent association with retention, so a single rank-boundary gap does not explain the observed variation in retention. Density is associated with retention in marginal analysis in all settings, and outside the top-10 neighborhood in all six settings tested with annulus density
(annulus ranks 11--30 partials 0.225 to 0.597), but contributes little additional explanatory power once energy and neighborhood contrast are included. Table~\ref{tab:joint}
summarizes the joint partials; Figure~\ref{fig:joint} plots them.

\begin{table}[t]
\centering
\small
\setlength{\tabcolsep}{3pt}
\begin{tabular}{lccc}
\toprule
Setting & density & energy & contrast \\
\midrule
CBOW & -0.06 & 0.57 & 0.17 \\
SG & 0.04 & 0.12 & 0.17 \\
RI & 0.12 & 0.20 & 0.13 \\
MiniLM & 0.07 & 0.22 & 0.26 \\
GTE & 0.09 & 0.25 & 0.25 \\
MPNet & 0.13 & 0.11 & 0.24 \\
Nomic PCA & -0.04 & 0.13 & 0.38 \\
Nomic native & 0.07 & 0.08 & 0.30 \\
\bottomrule
\end{tabular}
\caption{Joint partial correlations with retention at matched severity (static ret32, 384d ret64, 768d ret128), controlling boundary gap, contrast, energy (retained prefix-energy fraction for Nomic native), ID, hubness, frequency, and margin. After joint control, density partials are small across all settings (RI and MPNet remain Holm-significant but small); energy and contrast lead by regime. Joint-partial confidence intervals are shown in Figure~\ref{fig:joint}; boundary gap, ID, margin, and R-squared increments are reported in Table~\ref{tab:joint-rest}.}
\label{tab:joint}
\end{table}

The relationship with intrinsic dimensionality differs across representation regimes. On static CBOW and
Skip-Gram, density remains associated with retention after controlling
for ID (0.278 and 0.296 in the log1p-frequency specification), whereas the ID partial is near zero
($-0.048$ and $-0.124$), confirmed at a
scale-matched neighborhood (CBOW 0.341 at $k=1000$). On MiniLM and
GTE-small the pattern reverses: intrinsic dimensionality retains the
larger partial association ($-0.390$ MiniLM, $-0.386$ GTE, after controlling for
density), while density falls to 0.025
and 0.046.
Density remains independently associated on the static embeddings, whereas it largely vanishes after ID control on the modern encoders. The pairwise pattern completes across all eight settings: RI density 0.466 with ID at -0.224, MPNet density 0.091 with ID at -0.315, and Nomic density near zero on both paths with ID at -0.36 and -0.32. With the ID neighborhood at 10 instead of 50, MPNet density rises to 0.260, so the ID result is estimator-sensitive and reported as such. Marginal density partials by setting are plotted in Appendix Figure~\ref{fig:forest}.

\subsection{Robustness and interpretation}
\label{sec:robustness}

On MiniLM, changing the centering leaves the density partial unchanged (0.416 before and after); removing PC1 across spaces changes the measurement frame and the resulting cross-space partial is $-0.012$. Within a fixed frame, the estimates below use the same centering and frame throughout. Two checks locate the change: (1) within-PC1-bin and residualized estimates (0.472 and 0.469) show PC1 position alone does not explain density; (2) recomputation inside residual space gives 0.589. The near-zero cross-space estimate is consistent with the change in measurement frame. Within a fixed frame, denser neighborhoods keep ranks better in our tests.

Two qualifications remain. When retention is scored as rank displacement, the density partial is 0.186; when it is scored as preserved distance structure, the association is null, so part of the overlap result is specific to the overlap metric defined in Sec.~\ref{sec:method} as $R_{10}$. Matched-severity noise
retains a substantial but weaker association than PCA (0.282 vs.\ 0.416),
suggesting that generic neighborhood stability contributes to the association, while additional PCA-specific structure may remain. Controlling for full-dimension probe correctness changes nothing (0.415),
and within-frequency-decile analyses show no Simpson pattern on
any family. Retention falls monotonically with severity on MiniLM
(means 0.720, 0.626, 0.506, 0.383 at 256, 128, 64, 32 dimensions;
all adjacent drops positive); the density-rank association with the
256-to-32 drop is $-0.129$ (Spearman CI $[-0.193, -0.064]$; see Appendix B).
One interpretation is that retained energy captures how strongly an item's representation lies in the retained PCA subspace. Neighborhood contrast may capture a complementary source of stability: neighborhoods that are better separated from their local background have fewer close competitors after truncation. This pattern is consistent with density covarying with high-energy, high-contrast regions: density predicts retention marginally but contributes little additional information once those variables are included. Intrinsic dimensionality carries additional signal on modern encoders in pairwise models without energy (for example -0.390 MiniLM and -0.386 GTE-small), but in the joint model with energy controlled its partial falls to -0.09 MiniLM and -0.07 GTE (Table~\ref{tab:joint-rest}), and on MPNet it is sensitive to neighborhood size where density rises to 0.260 with ID at 10 instead of 50. Part of the pattern reflects generic neighborhood stability, which matched-severity noise reproduces in attenuated form, suggesting that additional PCA-specific structure may remain. The weak boundary-gap result suggests that neighborhood-level contrast is more informative than a single rank cutoff.

\section{Implications for Selective Allocation}
\label{sec:discussion}

We test whether assigning dimensions by predicted fragility beats uniform truncation at a matched mean budget. Median-split (48 above-median hubness, 16 rest), quartile (64 top quartile, 21 rest), and inverse (48 low-hubness, 16 high-hubness) allocations average 32 dimensions against uniform 32d; mixed-width pairs compare at the weaker word's width, with 20 random 48/16 splits as control. For retrieval, a hybrid routes below-median-density queries (about half) to 128d and the rest to 32d, averaging 80 dims per query against a uniform 80d comparator; each query is ranked independently with queries and corpus at the same width.
Median, quartile, and inverse per-word allocations show no improvement, with CIs covering zero and results within the 20-draw random-allocation distribution.
A retrieval hybrid routing low-density queries to 128d performs below uniform 80d (hybrid-minus-uniform $\Delta$nDCG difference $-0.062$ CI
$[-0.083, -0.042]$).

The aggregate retention-versus-dimension curve is concave, which favors uniform allocation under a shared-curve approximation. Selective allocation also faces a pairwise constraint: similarity quality depends on both endpoints, limiting the benefit of assigning extra dimensions to only one item. None of the tested selective budgets beats uniform budgets under the
tested operators. This negative result limits the tested operators.
Future methods may need to exploit structure beyond mean-preserving dimension reallocation, for example through reranking cascades.

\section{Conclusion}
\label{sec:conclusion}

Per-example neighborhood loss under PCA truncation is predictable from geometry. Retained subspace energy is positively associated with retention in all eight settings tested, as is neighborhood contrast, while density is associated with retention in marginal analysis but adds little unique information once they are controlled. Energy leads on CBOW and RI while contrast leads on SkipGram and the modern/Matryoshka settings, with GTE approximately tied. Density remains associated with retention outside the top-10 neighborhood in all six settings tested with annulus density. On SciFact, query density predicts gold retrieval degradation; on NFCorpus the primary 32d association does not replicate and a direct variance-restriction test leaves that null unresolved.

Compression evaluations should report subgroup or per-example degradation alongside aggregate means, since means can remain stable while damage concentrates in fragile neighborhoods. None of the selective allocation schemes we tested outperforms uniform truncation.

\section{Limitations}
\label{sec:limitations}

The downstream association is observed on only one of the two retrieval benchmarks.
The two retrieval benchmarks differ in both dataset and encoder, so the NFCorpus result cannot be attributed separately to either factor.
No adaptive method succeeded. We test
English only, a narrow set of 768d encoders (MPNet-base and Nomic v1.5), and one sentence corpus for geometry.
Lexical sources such as concreteness and polysemy remain untested,
as does bootstrap-retrain stability. Analogy evaluation is at floor,
so no analogy claims are made. The hubness control is less conclusive for GTE-small because its confidence interval extends farther from zero. Joint attribution depends on estimator choices (ID neighborhood size matters on MPNet). Geometry is a predictive correlate with stated limits, not an established cause.

\bibliography{references}

\appendix

\section{Additional Null Results}
\label{sec:appendix-nulls}

Each entry maps to its main-text section: adaptive allocations and
hybrid routing (Sec.~\ref{sec:discussion}), WordSim strata
(Sec.~\ref{sec:downstream}), and severity analysis
(Sec.~\ref{sec:robustness}).

\begin{table}[h]
\centering
\footnotesize
\setlength{\tabcolsep}{4pt}
\begin{tabular}{@{}p{3.0cm}p{4.35cm}@{}}
\toprule
Analysis & Result \\
\midrule
Median allocation & 95\% CI $[-0.059, +0.037]$; no improvement \\
Quartile allocation & 95\% CI $[-0.048, +0.025]$; no improvement \\
High-high pairs & $0.528$ vs $0.515$ (70 of 231 pairs); no statistically significant advantage \\
Inverse allocation & 95\% CI $[-0.054, +0.044]$; no improvement \\
Random allocation & Mean $-0.010$, range $[-0.045, +0.014]$; comparable \\
Retrieval hybrid & Hybrid-minus-uniform $\Delta$ $-0.062$, 95\% CI $[-0.083$, $-0.042]$; significantly worse \\
WordSim strata & Differences near $+0.03$ to $+0.04$ with CIs covering zero; no significant effect \\
Severity analysis & Monotone retention; Spearman $-0.129$ (CI $[-0.193$, $-0.064]$) significant in expected direction; prespecified positive criterion misspecified \\
\bottomrule
\end{tabular}
\caption{Null results with descriptive analysis names. CI = 95\% bootstrap confidence interval.}
\label{tab:nulls}
\end{table}

\section{Analysis Corrections}
\label{sec:appendix-corrections}

The severity analysis prespecified a positive directional criterion (Spearman at least $+0.50$). As coded, the 256-to-32 drop is positive when items lose more neighborhoods, so the scientifically expected association is negative and the preregistered sign was misspecified. The observed Spearman correlation of $-0.129$ (CI $[-0.193, -0.064]$) is significant in the scientifically expected direction. We retain the original criterion for transparency but do not use it to support the observed effect.

\section{Joint, Decoupled, ID, and Range-Restriction Results}
\label{sec:appendix-joint}

\begingroup
\raggedbottom
\setlength{\abovecaptionskip}{4pt}
\setlength{\belowcaptionskip}{4pt}

\begin{center}
\includegraphics[width=\columnwidth]{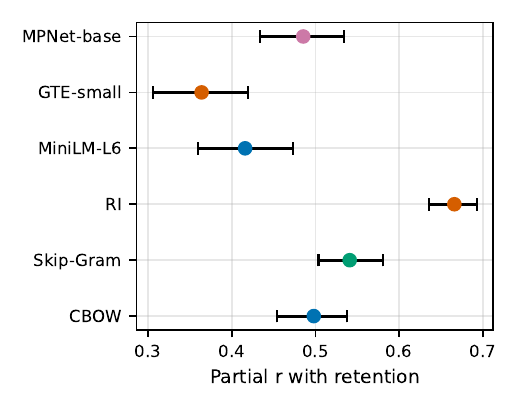}
\captionof{figure}{
Marginal density partial correlations with 95\% CIs across six PCA
settings at matched severity (static ret32, 384d ret64, 768d ret128;
frequency controlled). All six associations are positive, with
95\% confidence intervals excluding zero.
}
\label{fig:forest}
\end{center}

\medskip

\begin{center}
\small
\setlength{\tabcolsep}{3pt}
\begin{tabular}{lcccc}
\toprule
Setting & gap & ID & margin & $\Delta R^2$ \\
\midrule
CBOW & 0.02 & -0.09 & -0.04 & 0.002 \\
SG & 0.02 & -0.10 & -0.07 & 0.001 \\
RI & 0.05 & -0.12 & -0.03 & 0.007 \\
MiniLM & 0.06 & -0.09 & -0.10 & 0.003 \\
GTE & 0.04 & -0.07 & -0.06 & 0.005 \\
MPNet & 0.08 & -0.09 & -0.11 & 0.010 \\
Nomic PCA & 0.00 & -0.13 & -0.11 & 0.001 \\
Nomic native & 0.02 & -0.11 & -0.04 & 0.003 \\
\bottomrule
\end{tabular}
\captionof{table}{
Joint model companions: boundary gap, ID, and margin partials given
all other predictors, plus OLS R-squared increment from adding
density. Gap is near zero throughout; ID is negative and small;
density increments are 0.001 to 0.010.
}
\label{tab:joint-rest}
\end{center}

\medskip

\noindent\begin{minipage}{\columnwidth}
\centering
\small
\setlength{\tabcolsep}{2pt}
\begin{tabular}{@{}lcc@{}}
\toprule
Setting & density $|$ F+ID & ID $|$ F+D \\
\midrule
CBOW & 0.278 [0.229, 0.322] & -0.048 [-0.096, +0.000] \\
SkipGram & 0.296 [0.244, 0.346] & -0.124 [-0.170, -0.074] \\
RI & 0.466 [0.418, 0.510] & -0.224 [-0.276, -0.172] \\
MiniLM & 0.025 [-0.044, 0.094] & -0.390 [-0.447, -0.329] \\
GTE & 0.046 [-0.019, 0.108] & -0.386 [-0.444, -0.332] \\
MPNet & 0.091 [0.022, 0.157] & -0.315 [-0.377, -0.251] \\
Nomic PCA & 0.005 [-0.064, 0.071] & -0.358 [-0.415, -0.298] \\
Nomic native & 0.005 [-0.062, 0.073] & -0.323 [-0.384, -0.260] \\
\bottomrule
\end{tabular}
\captionof{table}{
Pairwise ID dual controls with the same Levina-Bickel estimator at
matched severity (F frequency, D density; log1p for CBOW/SkipGram,
raw for RI, rarity elsewhere). SkipGram ordering is sensitive to
this choice (raw 0.134/-0.217 vs log1p 0.296/-0.124) with the same
regime pattern either way. MPNet density rises to 0.260 with ID at
10 instead of 50, so the ID result is estimator-sensitive. Density
remains independently associated in static settings but is
substantially reduced after ID control in modern encoders; ID shows
the complementary pattern.
}
\label{tab:id-pairwise}

\end{minipage}

\medskip

\begin{center}
\small
\setlength{\tabcolsep}{3pt}
\begin{tabular}{lccc}
\toprule
Family & D10 & annulus 11--30 & annulus 11--50 \\
\midrule
CBOW & 0.498 & 0.457 & 0.449 \\
SG & 0.541 & 0.472 & 0.464 \\
RI & 0.666 & 0.597 & 0.582 \\
MiniLM & 0.416 & 0.327 & 0.313 \\
GTE & 0.364 & 0.225 & 0.201 \\
MPNet & 0.485 & 0.416 & 0.405 \\
\bottomrule
\end{tabular}
\captionof{table}{
Decoupled density partials with 95\% CIs excluding zero in every cell.
MPNet uses ret128; Sec.~6 reports ret64 value 0.517. Annulus predictors
exclude the top-10 set entirely yet retain 55 to 92 percent of the D10
association. Cross-representation MiniLM density versus GTE retention
is 0.184 CI [0.115, 0.251].
}
\label{tab:decoupled}
\end{center}

\medskip

\noindent\begin{minipage}{\columnwidth}
\centering
\small
\setlength{\tabcolsep}{3pt}

\begin{tabular}{lcccc}
\toprule
Band & n & SD & nDCG partial & recall partial \\
\midrule
A full & 300 & 0.086 & 0.249 & 0.266 \\
B      & 150 & 0.028 & 0.105 & 0.098 \\
C      & 128 & 0.023 & 0.085 & 0.022 \\
D      & 82  & 0.014 & 0.059 & 0.009 \\
\bottomrule
\end{tabular}

\captionof{table}{
SciFact 32d associations within restricted density bands.
Points attenuate monotonically but attenuation CIs cover zero
(Band C nDCG attenuation CI [-0.357, +0.030]), so range
restriction is unresolved as the NFCorpus explanation.
}
\label{tab:range}

\end{minipage}

\endgroup

\section{Reproducibility Details}
\label{sec:appendix-repro}

CPU-only Python 3.11 stack (numpy, scipy, sklearn, pandas, torch CPU,
sentence-transformers, datasets). Seeds: subset 7, with 5,000 bootstrap resamples for modern dual-controls. The neighbor module
was validated against brute-force search. All experiment scripts, per-row CSVs,
and JSONs will be released as supplementary material. Total
program compute under five hours wall clock.
The Nomic prompt prefix is the literal string \texttt{classification:\ } (with trailing space).

Prespecified replication criteria: a partial correlation counts as
replicated when its 95\% CI lower bound exceeds 0.20 at $n \ge 500$;
hubness counts as controlled when its density-controlled estimate lies
between $-0.10$ and $0.10$.

\end{document}